\documentclass[letterpaper, 10 pt, conference]{ieeeconf}

\IEEEoverridecommandlockouts
\usepackage{siunitx}
\usepackage{booktabs}
\usepackage{multirow}

\usepackage{array,multirow,graphicx}
\usepackage{color}
\usepackage{amsfonts}
\usepackage{amsmath}
\usepackage[dvipsnames]{xcolor}
\usepackage{wrapfig}
\usepackage[hidelinks]{hyperref}
\hypersetup{
    colorlinks=false,
    urlcolor=red,
    }
\usepackage{url}
\usepackage[backend=biber,
            hyperref=true,
            url=false,
            isbn=false,
            doi=false,
            backref=false,
            style=ieee,
            citestyle=numeric-comp,
            sorting=none,
            block=none]{biblatex}

\usepackage{comment}
\usepackage{float}
\usepackage[skip=2pt,font=scriptsize]{caption}

\usepackage[dvipsnames]{xcolor}
\usepackage{colortbl}

\newcommand{\ourmethod}{Exp-Force}

\title{\LARGE \bf
Exp-Force: Experience-Conditioned Pre-Grasp Force Selection with Vision-Language Models
}

\author{Siqi Shang*, Minchao Huang*, Bill Fan, and Lillian Chin%
\thanks{This work was partially supported by the Texas Robotics Industrial Affiliate Program.}
\thanks{*Equal contribution. All authors are with The University of Texas at Austin, TX, USA. Correspondence: {\tt\small siqi.shang@utexas.edu}}%
}

\begin{document}

\maketitle
\thispagestyle{empty}
\pagestyle{empty}

\begin{abstract}

Accurate pre-contact grasp force selection is critical for safe and reliable robotic manipulation. Adaptive controllers regulate force after contact but still require a reasonable initial estimate. Starting a grasp with too little force requires reactive adjustment, while starting a grasp with too high a force risks damaging fragile objects. This trade-off is particularly challenging for compliant grippers, whose contact mechanics are difficult to model analytically. We propose \ourmethod{}, an experience-conditioned framework that predicts the minimum feasible grasping force from a single RGB image. The method retrieves a small set of relevant prior grasping experiences and conditions a vision–language model on these examples for in-context inference, without analytic contact models or manually designed heuristics. On 129 object instances, Exp-Force achieves a best-case MAE of 0.43~\si{\newton}, reducing error by 72\% over zero-shot inference. In real-world tests on 30 unseen objects, it improves appropriate force selection rate from 63\% to 87\%. These results demonstrate that \ourmethod{} enables reliable and generalizable pre-grasp force selection by leveraging prior interaction experiences.

\end{abstract}

\section{INTRODUCTION}

Applying an appropriate amount of grasping force to an object is fundamental to successful robotic manipulation. Grasping force directly determines both grasp stability and object safety: too little force will cause the object to slip out of grasp, while too much force risks damaging fragile objects. 
Existing research on grasp force selection can be divided into \emph{pre-grasp force selection} and \emph{adaptive force control}. Pre-grasp methods select an initial force prior to contact based on object attributes or prior knowledge, while adaptive methods regulate force after contact using tactile feedback. Current research prefers the latter approach, as the force can be adjusted throughout the grasping process ~\cite{khamis2021realtime,schurmann2026affordable,yao2019design,qian2022compliant,yang2024contact,zhang2021adaptive,li2024multidofforce,Shang2026forte}. However, adaptive methods still require an initial guess of force. Starting with near-zero force reduces the risk of damaging delicate objects, but requires reliable slip detection to prevent failure on heavier or low-friction objects. Starting with a higher initial force increases grasp stability but risks damaging fragile objects. Pairing adaptive methods with a \emph{pre-grasp} force predictor can provide an appropriate ``just-enough'' force estimate before contact.


Generalizable pre-grasp force prediction remains an open problem. Classical analytical approaches derive grasping forces from contact mechanics and force-closure principles~\cite{bicchi2000robotic,cutkosky1989grasp}, but rely on precise prior knowledge that is typically unavailable for unseen objects (e.g., object geometry, friction, and contact conditions). Data-driven alternatives avoid explicit modeling by learning mappings from perceptual input to grasp force as part of a control policy~\cite{collins2023forcesight,kang2026learningforceregulate}. However, the generalization of such methods is inherently constrained by the diversity of object properties, interaction conditions, and gripper embodiments contained within the training dataset. The problem becomes even more challenging for compliant grippers, whose nonlinear and embodiment-dependent contact behaviors make accurate modeling particularly difficult~\cite{schurmann2026affordable,yao2019design,qian2022compliant}. 



\begin{figure}[t]
\centering
\includegraphics[width=\linewidth]{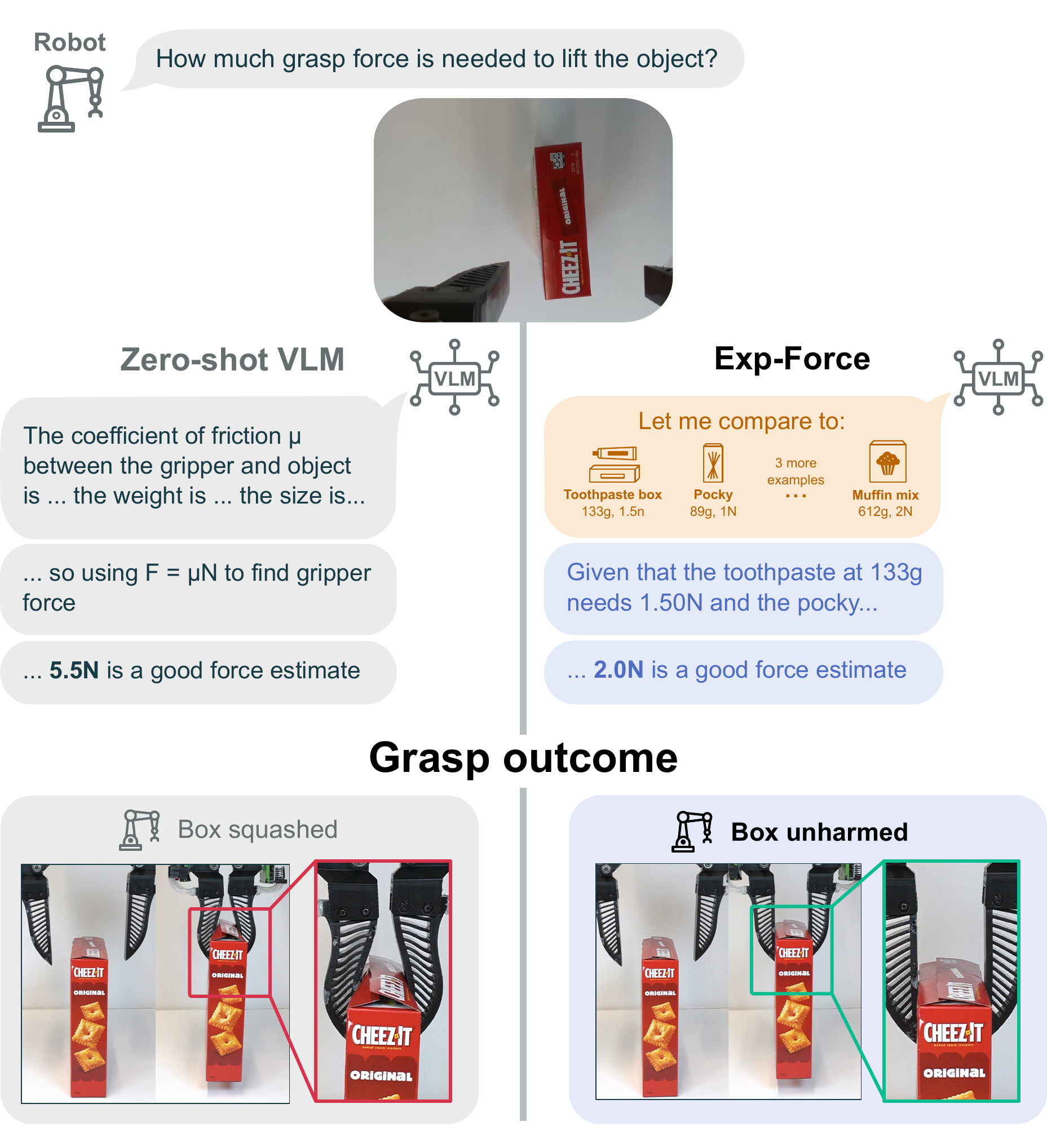}
\caption{\textbf{Overview}. \ourmethod{} uses VLMs to accurately predict appropriate grasp forces using a single object-image, prior to grasp contact. We do this by providing a VLM with a small set of related experiences for in-context learning. When grasping an opened Cheez-It box, a zero-shot VLM overestimates the grasp force by 2.75x, crushing the box, while \ourmethod{} can lift the box without damaging it.}
\vspace{-20pt}
\label{fig:overview}
\end{figure}

\begin{figure*}[!ht]
\centering
\includegraphics[width=\linewidth]{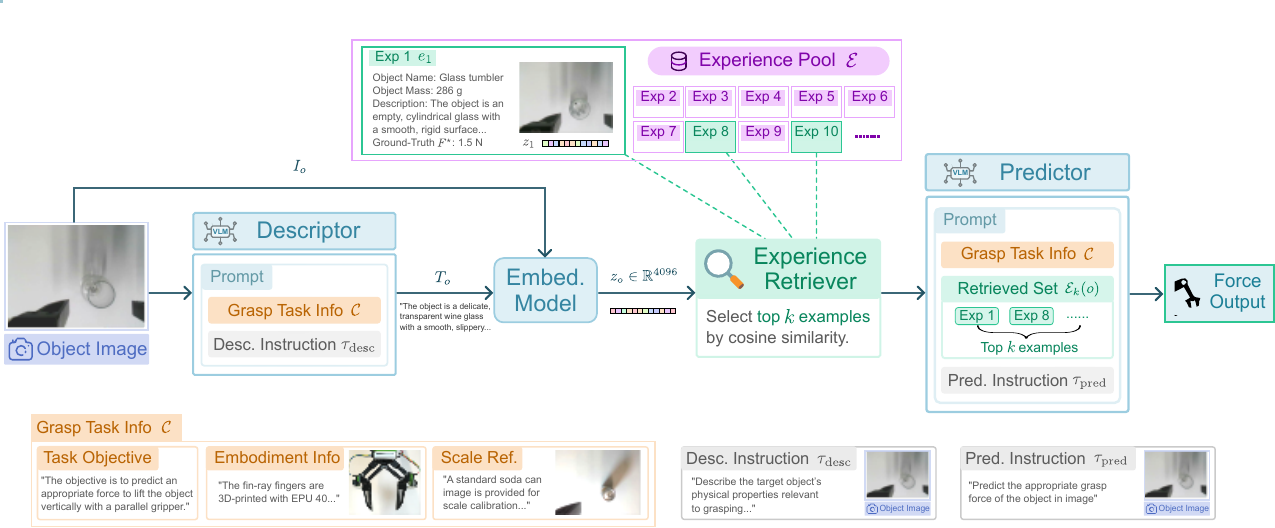}
\caption{\textbf{\ourmethod{} architecture.} A single target object image $I_o$ from the wrist camera is first fed into the descriptor VLM to generate a text description $T_{o}$ of the object's physical properties related to robot grasping. Together with the object image, the text description is fed into the embedding model to obtain the object embedding $z_{o}$. The embedding is used to perform a cosine-similarity search against the Experience Pool. The top-$k$ examples are retrieved and supplied to the predictor VLM as in-context examples for estimating the grasping force $\hat{F}_o$. Each example contains the object name, mass, description, image, and ground-truth $F^\star$.}
\label{fig:pipeline}
\vspace{-15pt}
\end{figure*}

We propose using foundation models to make pre-grasp force predictions, emulating how humans select grasp forces. Humans connect an object's appearance to prior interactions with visually and semantically similar objects. This process generates reliable force predictions by combining knowledge gained from previous grasps with current visual context~\cite{JohanssonFlanagan2009, JohanssonWestling1988, Paulun2019Material, Gallivan2014Action}. We adopt a similar \textit{experience-conditioned} approach to grasp force estimation using the knowledge encoded in visual-language models (VLMs). Large VLMs such as GPT-5.2 and Gemini-3.1-Pro encode human-like common-sense knowledge about object physical properties through massive image--text training data~\cite{shen2023physobjects,li2025physbench}. Prior work has shown that conditioning foundation models on retrieved examples through retrieval-augmented generation (RAG) can improve robotic planning and decision making by grounding model reasoning in relevant past experiences~\cite{Kagaya2024RAPRP,xie2024embodied,lan2025experience}.
Recent work has explored leveraging foundation models for forceful grasping by first inferring latent physical attributes (e.g., weight, friction, or compliance) from semantic descriptions and then combining these estimates with analytic formulations to compute grasping forces~\cite{deligrasp2024}. However, such approaches rely on explicit physical modeling assumptions and have been evaluated on limited object sets with rigid parallel grippers. 
In this paper, we propose \ourmethod, an experience-conditioned framework that predicts the appropriate grasping force for an object (Fig.~\ref{fig:overview}). Given a single RGB image of a novel object, \ourmethod{} predicts the appropriate grasping force by generating a task description with a VLM, retrieving relevant experiences from a small pool of prior grasping experiences, and incorporating those examples into the prompt of a force predictor VLM. Our approach avoids injecting human-designed heuristics or analytic modeling into the prediction process, but instead maps wrist-camera object images directly to grasping force estimates. Importantly, the framework is platform-agnostic and does not depend on a specific gripper embodiment.

We demonstrate through offline and real-world experiments that \ourmethod{} enables data-efficient and generalizable grasping force estimation in practical robotic settings. Using a modest experience pool of only 129 object grasps ($\approx$ 1 hour of real robot data), our method achieves accurate and stable force predictions with a best-case cross-validation mean absolute error of $0.43\,\si{\newton}$. The strong predictive performance of our framework suggests that the complex interaction physics between compliant grippers and diverse objects can be implicitly captured by VLMs when grounded in a small set of relevant interaction examples.

In summary, our contributions are:

\begin{enumerate}

    \item To the best of our knowledge, this is the first work to leverage retrieval-augmented VLMs for estimating appropriate grasping forces in robotic manipulation.

    \item We demonstrate the effectiveness of \ourmethod \ with offline evaluation on a dataset of 129 object instances. \ourmethod \ achieves a force prediction error of $0.43$\,\si{\newton}, outperforming prediction of zero-shot VLMs by 72\%

    \item We show that accurate force estimation needs only a small ($k=6$) number of relevant examples, highlighting the in-context sample efficiency of \ourmethod.

    \item We validate the generalization capability of \ourmethod{} through real-world robotic experiments across diverse and unseen objects using compliant robot fingers.

    \item We release our dataset and implementation at 
    
    {\color{orange} \url{http://expforcesubmission.github.io/Exp-Force-Website/}}

\end{enumerate}

\section{Methods}




Our objective in this work is to predict the approrpiate grasping force scalar $F^\star$ when given a single wrist-camera image $I_o$ of an object instance $o$. We define $F^\star$ as the minimum grasping force needed  to lift the object without slip. Grasp force is defined as the sum of the contact normal forces of two parallel gripper fingers against the object.


To achieve this objective, we formulate force prediction as an experience-conditioned inference problem: a small set of relevant prior grasping experiences is retrieved and provided to a VLM as contextual examples to estimate $F^\star$ for a query object $o$ (Fig.~\ref{fig:pipeline}).
Our pipeline consists of three main components: \textit{(A) Object description generation}, \textit{(B) Experience retrieval}, and \textit{(C) Experience-conditioned force inference}. Steps A and C are VLM-based and use structured prompts to ensure that inference is performed under well-defined task and embodiment assumptions. 


\subsection{Object Description Generation}
\label{sec:approach_description}
The first step generates a textual description of the query image to explicitly identify object attributes that are relevant to grasping force estimation. We obtain this description by constructing the following prompt for a descriptor VLM: a shared grasp-task information $\mathcal{C}$, a task-specific description instruction $\tau_{\mathrm{desc}}$, and the input query image $I_o$. This construction standardizes information about the grasp task across different object instances. 

The grasp-task information $\mathcal{C}$ provides contextual information to the VLM. It specifies (1) the task objective to find $F^\star$; (2) embodiment information about the gripper, such as geometry, structure, material, and a side-view image; and (3) a visual scale reference image captured under identical camera settings to provide metric size cues. he shared context $\mathcal{C}$ is provided to both the descriptor and predictor (Sec.~\ref{sec:approach_prediction}) to ensure consistent physical assumptions and manipulation constraints. The instruction $\tau_{\mathrm{desc}}$ explicitly directs the model to describe object properties relevant to $\mathcal{C}$. The description instruction $\tau_{\mathrm{desc}}$ specifies that the model should characterize visual and semantic attributes relevant to grasping force estimation. Importantly, no analytic force models or manually designed heuristics are embedded in the instruction prompt. 

Formally, let $\Pi_{\mathrm{desc}}(\cdot)$ be the prompt construction function that formats $\mathcal{C}, \tau_{\mathrm{desc}}, I_o$ into a structured multi-modal prompt for the descriptor module.  The descriptor VLM $g_{\mathrm{desc}}$ then processes the constructed prompt to generate a textual description $T_o$ of the object's physical properties:
\begin{equation*}
T_o = g_{\mathrm{desc}}\!\left(
\Pi_{\mathrm{desc}}(
\mathcal{C},
\tau_{\mathrm{desc}},
I_o
)
\right).
\end{equation*}

In practice, we find that the output $T_o$, offers several semantically relevant characteristics about the target object $o$. Example descriptions that the VLM returns include size, shape, apparent rigidity or compliance, surface characteristics, and structural cues (ex. container-like vs. solid objects). This demonstrates that  $T_o$ serves as a semantic representation that can support similarity-based retrieval in the next stage.

\begin{figure*}[!ht]
\centering
\includegraphics[width=\linewidth]{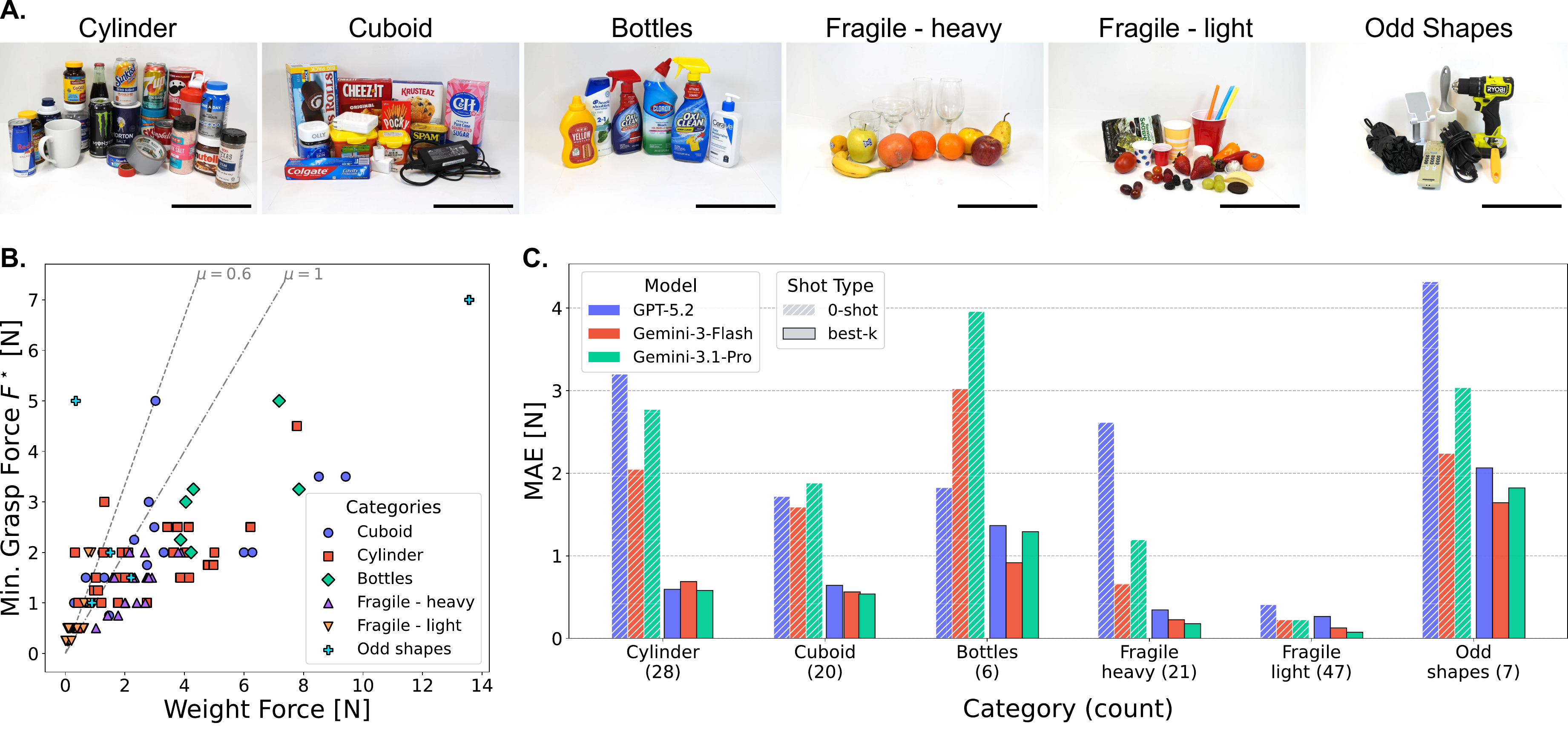}
\caption{\textbf{Evaluation objects, categories, and model performance across categories.} \textbf{\textit{(A)}} Representative objects used in our experiments, categorized by geometry, fragility, and weight. All scale bars are 20cm. \textbf{\textit{(B)}} Objects weight versus minimum feasible grasping force $F^\star$. Dashed lines indicate bounds for the coefficient of friction $\mu$ that the zero-shot model uses in its force estimation. \textbf{\textit{(C)}} Comparison of the Mean Absolute Error between zero-shot versus \ourmethod{} in each of the six object categories, across three vision-language models (GPT 5.2, Gemini-3-Flash, and Gemini-3.1-Pro). For each model, we chose the best $k$ value from our in-context sample efficiency ablation (Sec.~\ref{sec:k-sweep}). The total count of objects within each category is also reported.}


\label{fig:objects}
\vspace{-15pt}
\end{figure*}

\subsection{Experience Retrieval}
\label{sec:approach_retrieval}
Next, we use the generated object description $T_o$ and the object image $I_o$ to retrieve similar prior experiences to estimate $F^\star$. Similarity is computed by embedding both the query object and prior experiences into a shared multimodal embedding space. The multimodal embedding is computed using a pretrained vision--language embedding model $\phi$:
\begin{equation*}
z_o = \phi(I_o, T_o),
\end{equation*}
where $z_o \in \mathbb{R}^d$ denotes the embedding of the query object $o$, capturing both visual and semantic features, and $d$ is the dimensionality of the embedding space. In this work, we adopt the Qwen3-VL-Embedding-8B model~\cite{qwen3vlembedding} as the embedding function $\phi$, for which $d=4096$. We choose Qwen3 as it is specifically optimized for multimodal retrieval tasks, mapping both the visual features of the query image $I_o$ and the semantic description $T_o$ into a unified vector space.

We then compare $z_o$ with the embeddings of prior experiences. We maintain an experience pool:
\begin{equation*}
\mathcal{E} = \{ (n_i, m_i, T_i, I_i, F_i^\star) \}_{i=1}^{N},
\end{equation*}
where each experience $e_i$ consists of the object's name $n_i$, mass $m_i$, text description $T_i$, pre-grasp image $I_i$ and the ground truth $F_i^\star$ obtained from prior robotic interaction (Section.~\ref{section:experience_pool}). We use $\phi$ to compute the embedding of each stored element in this experience pool ($z_i$) and calculate the cosine similarity between $z_i$ and the query embedding $z_o$:
\begin{equation*}
\mathrm{sim}(z_o, z_i) = 
\frac{ z_o^\top z_i }
     { \| z_o \|_2 \, \| z_i \|_2 }.
\end{equation*}

We rank experiences in $\mathcal{E}$ by $\mathrm{sim}(z_o, z_i)$ and select the $k$ highest ones as our retrieved experience set:
\[
\mathcal{E}_k(o)
= \{ (n_i, m_i, T_i, I_i, F_i^\star) \mid i \in \mathcal{I}_k(o) \},
\]
where $\mathcal{I}_k(o)$ contains the indices of the top-$k$ similar embeddings. $\mathcal{E}_k(o)$ serves as the experience context that will be used to inform the force predictor module.

\subsection{Experience-Conditioned Force Inference}
\label{sec:approach_prediction}
In the final stage, we condition a predictor VLM on the retrieved experiences $\mathcal{E}_k(o)$ to estimate the appropriate grasping force $F^\star$ through in-context inference. We construct a predictor prompt that integrates four components in sequence: the shared grasp-task information $\mathcal{C}$ defined in Sec.~\ref{sec:approach_description}, the selected prior interactions $\mathcal{E}_k(o)$, a task-specific force prediction instruction $\tau_{\mathrm{pred}}$ that directs the model to infer $F^\star$, and the query image $I_o$. 
The predictor VLM $g_{\mathrm{pred}}$ then processes this prompt to produce an experience-conditioned estimation of the scalar minimum feasible grasping force $\hat{F}_o$:
\begin{equation*}
\hat{F}_o =
g_{\mathrm{pred}}\!\left(
\Pi_{\mathrm{pred}}(
\mathcal{C},
\mathcal{E}_k(o),
\tau_{\mathrm{pred}},
I_o,
)
\right).
\end{equation*}

In this paper, we instantiate both the descriptor $g_{\mathrm{desc}}$ and predictor $g_{\mathrm{pred}}$ using three state-of-the-art vision–language models: GPT-5.2, Gemini-3-Flash, and Gemini-3.1-Pro. We will compare these different models in Sec.~\ref{sec:offlineResults} to examine robustness across different VLMs.

Importantly, no analytic contact models, friction equations, or manually designed force heuristics are incorporated into the prediction prompt. This means that the whole \ourmethod \ pipeline operates end-to-end from visual observation alone, avoiding the need for explicit physical modeling.


When $k = 0$, no prior experiences are provided in the prompt, and the formulation reduces to zero-shot inference:
\begin{equation*}
\hat{F}_o^{(0)} =
g_{\mathrm{pred}}\!\left(
\Pi_{\mathrm{pred}}(
\mathcal{C},
\tau_{\mathrm{pred}},
I_o
)
\right).
\end{equation*}
This special case enables direct comparison between zero-shot VLMs and \ourmethod, isolating the contribution of experience conditioning.

\section{Experimental Setup}

To evaluate the performance of \ourmethod{} on predicting the target grasping force $F^\star$, we compare \ourmethod{} against baselines with offline and real-world experiments. We structure our evaluation around the following questions:
\begin{itemize}
    \item Can \ourmethod{} enable accurate prediction of $F^\star$ of diverse unseen objects?
    \item How much does each component of \ourmethod{} contribute to the force prediction performance?
    \item How many in-context examples do we need to accurately predict the target force $F^\star$?
\end{itemize}

\subsection{Experience Pool Data Collection} \label{section:experience_pool}
Accurate evaluation of \ourmethod{} requires both testing across diverse objects and accurately measuring the true minimum feasible grasping force $F^\star$. This section describes our choice of objects and method for measuring $F^\star$.

We first assembled a set of 129 objects categorized into 6 semantic categories. Objects were selected by combining household items from a grocery store with items from the YCB dataset \cite{Calli_2017_YCB} to ensure a variety in geometry, weight, fragility, and surface texture (Fig.~\ref{fig:objects}A). Object categories were created by first isolating fragile objects and classifying them as Fragile–light'' or Fragile–heavy’’ based on whether their mass was below or above 100,\si{\gram}. The remaining objects were classified by grasp geometry. We found that most durable household items could be classified as either ``Cuboid'' or ``Cylinder''. Of outliers, ``Bottles'' were a salient category. Many liquid products are packaged in slender, slippery, and heavy bottles. All remaining outliers were classified as "Odd shapes".


For each object, we collect real-time grasping force and slip measurements using a pair of FORTE tactile fin-ray fingers \cite{Shang2026forte} on a linkage-based gripper \cite{seo2024legato}, as well as an object RGB image (640 x 480 resolution) collected by a RealSense D435i wrist camera.  The force estimation runs at 100~Hz with $<$\,10~\si{\milli\second} latency, while slip detection runs at 500~Hz with $<$\,100~\si{\milli\second} latency. This wrist camera captures a top-down view image before each data collection trial to serve as a pre-grasp image $I_o$ for our experience pool. The entire end-effector assembly is attached to a Franka Emika Panda robot arm to generate a vertical lifting motion.

To accurately measure ground-truth $F^\star$ for each object, the gripper uses adaptive force control. The gripper first applies a small grasping force of 0.25~\si{\newton} to the object and begins to lift the object by 5\,\si{cm} at 0.5\,\si{cm}/\si{s}. During this 10 second period, for each slip event detected, the gripper tightens its grasp by $\approx$2\,\si{mm} (inter-finger distance), which increases the grasp force. Since grasp force is only increased until there is no longer slippage, the final force measurement accurately measures the minimum grasp force $F^\star$ required for lifting the object. We performed three lifting trials for each object, and selected the median value. Finally, we ceiling-round the value to 0.25~\si{\newton} increments to account for sensor noise and the grasp force estimator's RMSE ($\approx$ 0.2~\si{N}). Before applying this approach for finding $F^\star$ to all 129 objects, we verified that $F^\star$ was accurate across 20 objects through robot teleoperation by confirming that grasp forces under $F^\star$ were insufficient to lift the object. While scaling this method to the full dataset, we found that adaptive force control failed for 10 objects (Pocky, drill, lotion bottle, Clorox toilet bleach, Oxiclean 21.5oz, mustard, sugar, bean paste 28oz, muffin mix, black AC adapter). We established the ground truth $F^\star$ for these objects through robot teleoperation instead.




Figure~\ref{fig:objects}(b) plots the measured $F^\star$ in relation to object weight. We see our semantic categories correspond to distinct regions in the plot: fragile items cluster to the bottom left, having low weight and requiring low force; bottles cluster in the center being heavy and requiring moderately high forces; finally, odd shapes occupy extremes, such as the slippery peeler at the top left, and the heavy drill at the top right. This plot also enables us to compare the friction between the gripper and object against analytic friction models. Under a classical Coulomb friction model, $\lvert \textbf{F}_\textrm{weight}\rvert = \mu F_\textrm{grasp}$. Thus, different $\mu$ values determine the slopes of lines through the origin. In Fig.~\ref{fig:objects}(b) we plot lines corresponding to the typical range of $\mu$ values selected by our zero-shot models. Since most objects lie below the $\mu = 1$ line, this indicates the gripper's effective $\mu$ is much higher than 1 on most objects. As we later discuss in Section \ref{sec:experience_impact}, this is the primary cause for our zero-shot VLMs consistently overestimating the appropriate grasp force $F^\star$.

\subsection{Offline Experiments}
After data collection of the experience pool, we evaluate how well \ourmethod{} fares in pure force prediction on the dataset. We compare \ourmethod{} against three types of baselines: (1) zero-shot VLMs to isolate the performance benefit of retrieving prior interaction examples, (2) standard supervised learning baselines, and (3) ablations to assess the contributions of the predictor module, experience retriever, and embodiment information provided in the prompts.

\begin{figure*}[!ht]
\centering
\includegraphics[width=\linewidth]{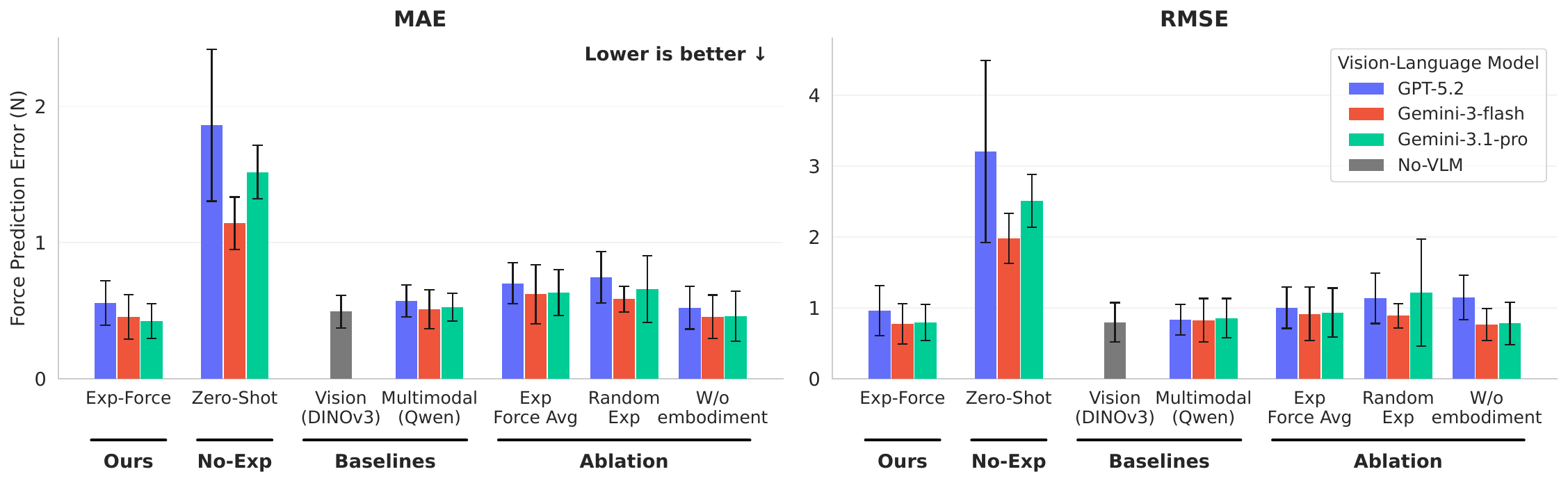}
\caption{Comparison of force prediction errors across three vision–language models (GPT-5.2, Gemini-3-flash, and Gemini-3.1-pro). Lower values correspond to more accurate force estimation. For the multimodal baseline, we utilize Qwen3-VL-Embedding (denoted as Qwen). For each model, distribution bars summarize the error statistics over five-fold cross-validation, where the thin whiskers denote one standard deviation. 
}
\label{fig:mae}
\vspace{-15pt}
\end{figure*}

For Category (2), we compare \ourmethod{} against a representative pretrained vision encoder (DINOv3-Large~\cite{simeoni2025dinov3}) and a representative multi-modal encoder (the previously-used Qwen3-VL-Embedding-8B~\cite{qwen3vlembedding} from Sec.~\ref{sec:approach_retrieval}) to incorporate object descriptions $T_o$. In both cases, the encoder parameters are frozen, and a two-layer MLP head is trained on top of the extracted features to predict the $F^\star$. We perform an extensive hyperparameter sweep over hidden dimension, dropout rate, learning rate, and weight decay (512 configurations per model). The results reported in Fig.~\ref{fig:mae} correspond to the best-performing checkpoints from the hyperparameter sweep on the validation splits, providing an optimistic estimate of achievable performance.

For Category (3), first we evaluated a simple baseline that averages the ground-truth forces $F_i^\star$ of the top-$k$ retrieved examples (\textit{Exp Force Avg}). Second, we replace the retrieved set $\mathcal{E}_k(o)$ with $k$ randomly sampled experiences (\textit{Random Exp}). Lastly, we remove the gripper embodiment information from both the descriptor and predictor prompt (\textit{w/o embodiment}).

Beyond the study of model components, we conduct an additional ablation study on the number of selected samples $k$ to understand how the size of the retrieved set $\mathcal{E}_k(o)$ affects prediction performance. Specifically, we vary $k$ and measure the resulting force estimation accuracy. This experiment characterizes how performance scales with the quantity of conditioning examples and provides insight into the in-context sample efficiency of \ourmethod{}.


For all offline experiments, we evaluate Exp-Force using 5-fold cross-validation across the entire collected dataset of 129 object instances to assess generalization to unseen objects. In each fold, the available objects are partitioned into a set of query objects for force prediction and a remaining set used to construct the experience pool $\mathcal{E}$. Across the five folds, every object is evaluated as a query, ensuring comprehensive coverage of the dataset while guaranteeing that no query instance ever appears in its corresponding retrieved set.

\begin{figure}[t]
\centering
\includegraphics[width=\columnwidth]{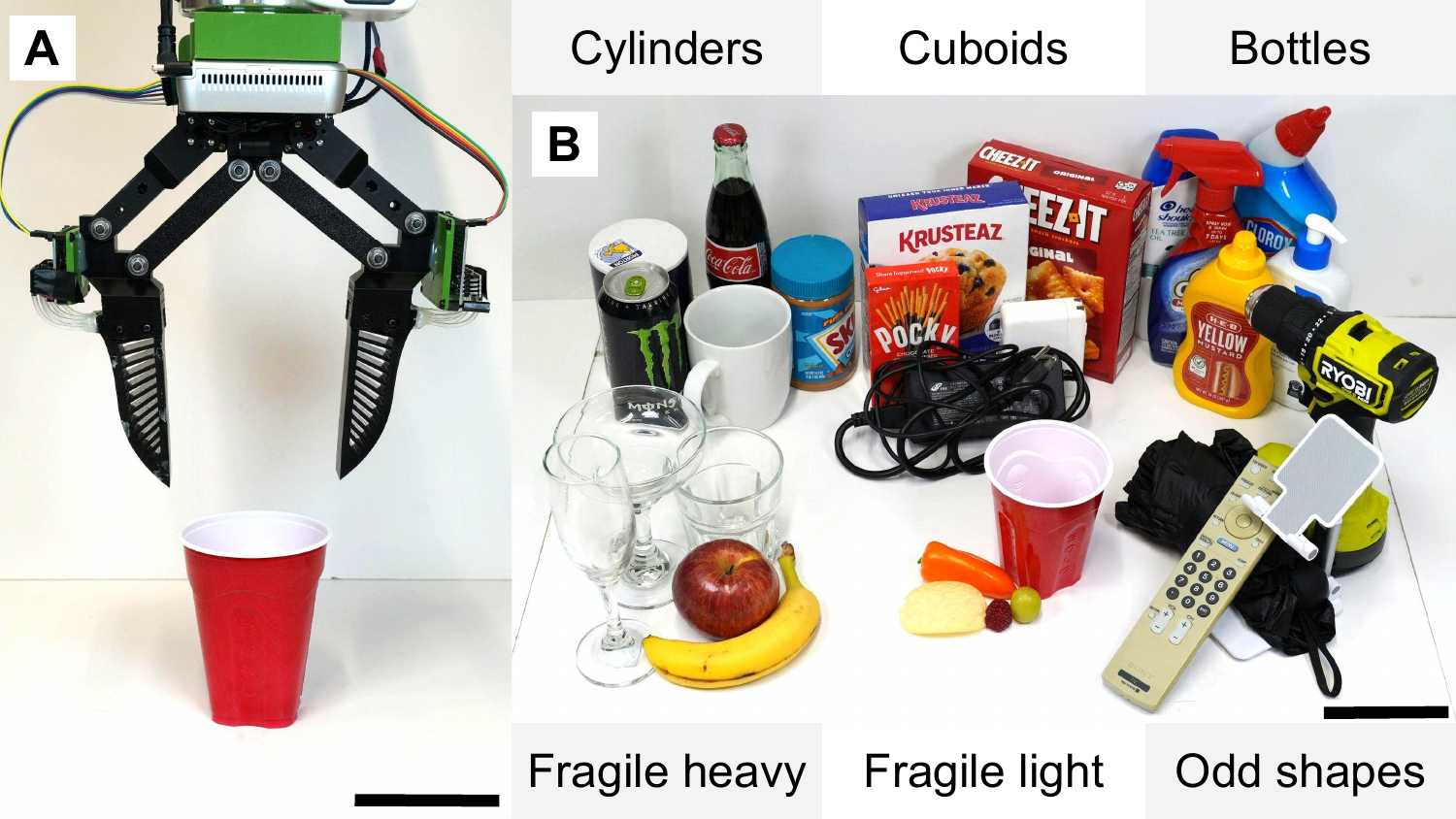}
\caption{\textbf{Real-world experiment setup.}
\textit{(A)} The compliant tactile two-finger gripper with a wrist-mounted camera used for pre-grasp image capture and force execution.
\textit{(B)} Five selected objects from each of the six categories. All scale bars are 10cm. }
\label{fig:real_setup}
\vspace{-15pt}
\end{figure}

\subsection{Real World Experiments}

To evaluate \ourmethod{} in the real world, we grasp objects with the force estimation $\hat{F}_o$ given by \ourmethod{} and the zero-shot VLM baseline (Fig.~\ref{fig:real_setup}A). We perform our real-world evaluations on the same grasping platform we used to collect the experience pool dataset (Sec.~\ref{section:experience_pool}). We compare the performance of \ourmethod{} and zero-shot VLM over 150 grasps: 5 grasp trials each for 30 objects. The thirty objects were chosen by selecting five objects with different sizes, textures, and weights from each of the six categories described in Sec.~\ref{section:experience_pool}, and are shown in Fig. \ref{fig:real_setup}B. We chose Gemini-3-flash as the selected VLM based on our results in Sec.~\ref{sec:offlineResults}. Although our offline experiments showed that Gemini-3-pro is our most performant model, we found Gemini-3-pro exhibited inconsistent response latency under high API load during our experiments.

In each grasp trial, we first perturb the pose of the target object, and then take an image with the robot wrist camera from a consistent end-effector pose. We then use this image to query both a zero-shot VLM and \ourmethod{} ($k=7$) for their pre-grasp force estimates. Finally, for each of the two force outputs, we manually center the target object under the gripper and perform a grasp. To simplify the real-world experiments, trials whose outcomes could be inferred from previous trials were not repeated. For example, if an object was crushed at 5\,\si{\newton}, we did not perform further trials on the same object with larger forces. Grasp forces with inconsistent grasp outcomes were re-run three additional times, with the more common outcome taken as the final outcome.

\begin{figure}
\centering
\includegraphics[width=\linewidth]{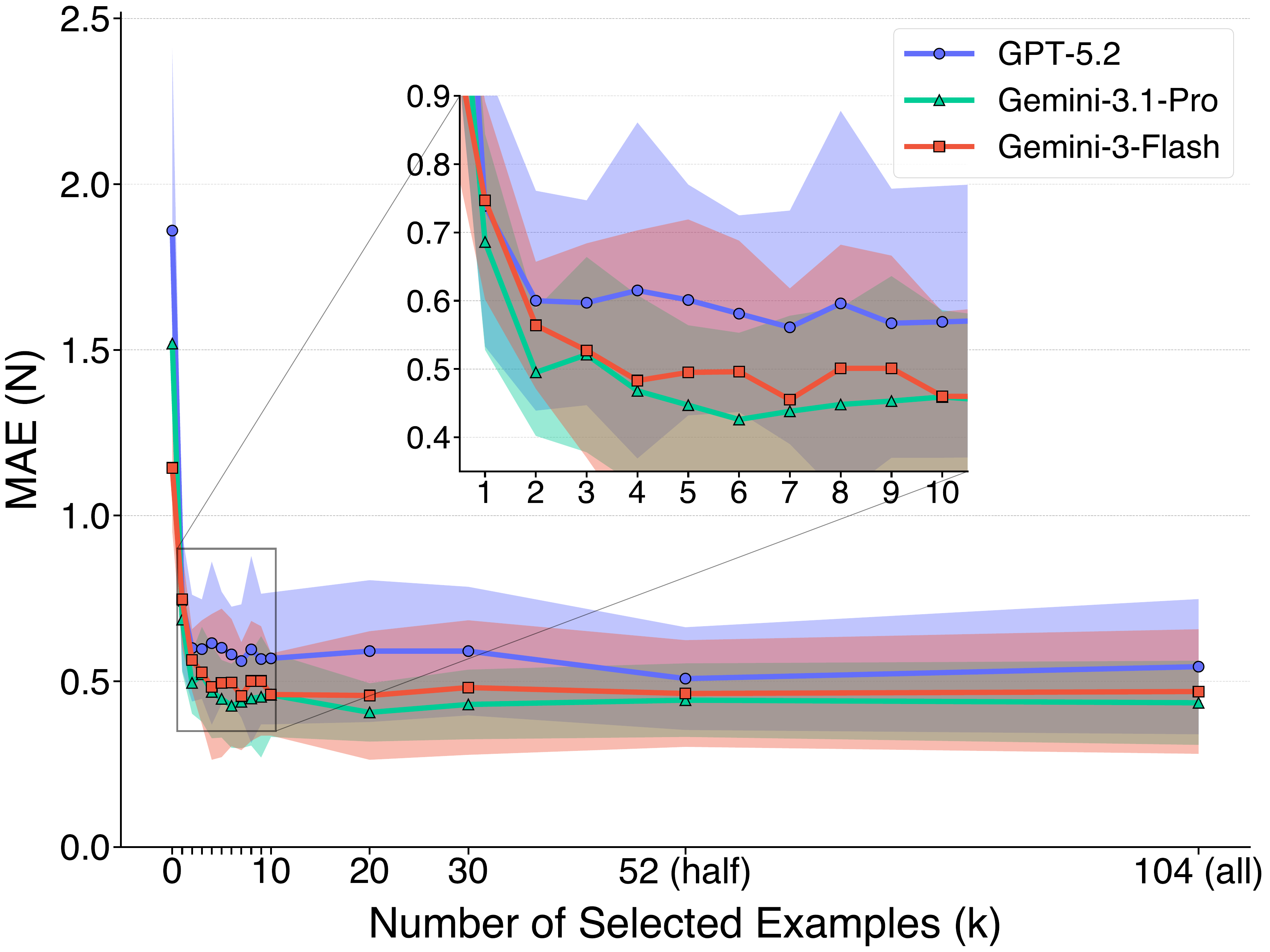}
\caption{Effect of the retrieved example size ($k$) on force estimation error. Our evaluation reveals that \ourmethod{} exhibits remarkable in-context sample efficiency. The plot demonstrates that, across all predictor VLMs, the Mean Absolute Error (MAE) rapidly converges as the model switches from zero-shot inference ($k=0$) to experience-conditioned inference. Performance plateaus early, typically between $k=5$ and $k=10$, highlighting the high sample efficiency of \ourmethod{}. Shaded regions denote the standard deviation across cross-validation folds, illustrating that the Gemini models maintain superior estimation stability and tighter error bounds compared to GPT-5.2.}
\label{fig:mae_vs_k}
\vspace{-20pt}
\end{figure}

We categorize the grasp outcome as Appropriate, Overestimate, or Insufficient. A prescribed force is Insufficient if a grasp cannot be stably maintained, manifesting as either an initial failure to pick up the object, or the object slipping out of the grippers within one minute of lifting. If a grasp is successful, its force is categorized as an Overestimate if the commanded force is greater than $3\times$ the ground-truth $F^\star$, or by $4$N. Otherwise, if a grasp is successful and the force is not overestimated, then it is classified as Appropriate. We choose to identify Overestimates as a standalone category because excessive grasp force is undesirable even if it does not lead to damage to the grasped object. Excessive force induces unnecessary wear on the gripper and tactile sensors. Also, fragile objects may appear deceptively non-fragile, such as the opened Cheez-It box in Fig.~\ref{fig:overview}. 

\section{Offline Experiment Results}
\label{sec:offlineResults}

\subsection{Comparison to Zero-Shot VLMs} \label{sec:experience_impact}


Overall, our results show that incorporating selected prior examples into the prompt substantially improves force prediction accuracy compared to zero-shot inference. \ourmethod{} consistently reduces error across all three tested VLMs and all six object categories (Fig.~\ref{fig:objects}C, \ref{fig:mae}-No-Exp). Specifically, the mean absolute error (MAE) decreases from 1.86\,\si{\newton} to 0.56\,\si{\newton} for GPT-5.2, from 1.14\,\si{\newton} to 0.46\,\si{\newton} for Gemini-3-Flash, and from 1.52\,\si{\newton} to 0.43\,\si{\newton} for Gemini-3.1-Pro. The consistently higher root-mean squared error (RMSE) compared to MAE indicates the presence of larger errors on objects requiring higher $F^\star$. These results show that incorporating selected prior examples into the prompt substantially improves force prediction accuracy compared to zero-shot VLMs.

To understand the mechanisms driving the accuracy improvement over zero-shot VLMs, we qualitatively analyzed the reasoning traces from the predictors $g_{\mathrm{pred}}$. While the VLMs usually successfully recognized the objects, the zero-shot $g_{\mathrm{pred}}$ consistently defaults to explicit analytical modeling. For example, when predicting $F^\star$ for a 236\,\si{\gram} Cheez-It crackers box (Fig.~\ref{fig:overview}), the $g_{\mathrm{pred}}$ recognizes the object but hallucinates unobservable physical parameters (e.g., guessing a friction coefficient of $\mu \approx 0.6$) and applies a rigid-body Coulomb friction model ($F_{\text{grip}} \ge W / \mu$). This analytical approach is fundamentally brittle as the simplified modeling equations fail to capture the deformation and form-closure of compliant fin-ray fingers (Sec.~\ref{section:experience_pool}). As a result, the zero-shot VLMs overestimate the required force to be 5.5\,\si{\newton}. 

\ourmethod{} circumvents this failure mode by allowing $g_{\mathrm{pred}}$ to act as a semantic-physical interpolator rather than a flawed physics engine. In the Cheez-It box example, \ourmethod{}'s $g_{\mathrm{pred}}$
specifically cited the examples of a muffin box at 612g and a toothpaste box of 133g to make the prediction, as they share \textit{``similar flat, smooth cardboard properties''} to the target Cheez-It cracker box. It also implicitly reasons about complex contact mechanics, stating that the force \textit{``accounts for the load and ensures the fin-ray fingers conform well enough without
causing structural damage to the box.''} These results indicate that VLMs can implicitly account for hard-to-model physical interactions when grounded by a relevant set of real-world examples.

\subsection{Comparison to Supervised Baselines}
\ourmethod{} also outperforms the standard supervised learning baselines of DINOv3-Large and Qwen3-VL-Embedding-8B (Fig.~\ref{fig:mae}-Baselines). 
While these models achieve reasonable mean absolute errors ($0.50$\,\si{\newton} and $0.51$\,\si{\newton}, respectively, with the latter using $T_o$ from Gemini-3-Flash), 
\ourmethod{} paired with Gemini-3.1-Pro or Gemini-3-Flash outperforms them by $14\%$ and $8\%$, respectively. This suggests that conditioning the VLM on retrieved interaction examples for in-context inference enables more adaptive force estimation than regression based solely on distances in a frozen embedding space.

This advantage over supervised baselines reveals a key difference in task formulation. Standard data-driven approaches attempt to learn a global mapping from observation directly to a continuous grasping force value, which can be prone to overfitting given our modest dataset of 129 object instances. This is because the $F^\star$ varies substantially across diverse objects as gripper-surface interactions differ. \ourmethod{} avoids this overfitting by reformulating the problem from absolute continuous regression to local comparative reasoning. 


\subsection{Ablation Studies}
We isolate the contributions of each component of \ourmethod{} for ablation studies (Fig.~\ref{fig:mae}-Ablations). \ourmethod{} achieves lower errors than the \textit{Exp Force Avg} baseline across all three selected VLMs, demonstrating that $g_{\mathrm{pred}}$ is actively reasoning about different object properties, rather than simply interpolating the aggregated examples. The \textit{Random Exp} baseline also performs uniformly worse across all VLMs, confirming that the performance of \ourmethod{} relies on the specific semantic and physical relevance of the retrieved embeddings $z_i$, not just the structural format of the prompt.

Surprisingly, the \textit{w/o embodiment} baseline performs only marginally worse than \ourmethod{}. Across all three tested VLMs, the largest increase in mean absolute error when removing the embodiment information is only $0.034\,\si{\newton}$. This suggests that the predictor  $g_{\mathrm{pred}}$ implicitly deduces the embodiment constraints through in-context inference directly from the retrieved experience set $\mathcal{E}_k(o)$; i.e., the in-context examples effectively act as an implicit, data-driven physical model of the hardware. The necessity for explicit text-based embodiment hints is diminished, because the ground-truth forces $F_i^\star$ provided in the examples inherently encapsulate the complex contact mechanics of the specific gripper.

\subsection{In-context Sample Efficiency and Diminishing Returns} \label{sec:k-sweep}
To understand how many prior experiences are needed for effective in-context inference, we vary the number of retrieved examples $k$ and measure the force prediction error. Our evaluation reveals that \ourmethod{} exhibits remarkable in-context sample efficiency (Fig.~\ref{fig:mae_vs_k}). The rapid reduction in estimation error from zero-shot ($k=0$) to experience-conditioned inference suggests a fundamental shift in the predictor VLM's reasoning mechanism when anchored by the retrieved set $\mathcal{E}_k(o)$. Performance plateaus early, typically between $k=5$ and $k=10$. Within this optimal data-efficient window, Gemini-3.1-Pro achieves the lowest MAE ($0.43\,\pm\,0.13$\,\si{\newton} at $k=6$), closely followed by Gemini-3-Flash ($0.46\,\pm\,0.16$\,\si{\newton} at $k=7$) and GPT-5.2 ($0.56\,\pm\,0.16$\,\si{\newton} at $k=7$). Expanding the retrieved experience set beyond $k=10$ yields diminishing returns. As a result, the models require only a small, highly relevant set of semantic neighbors to accurately anchor their estimations of $F^*$. This diminishing return implies that once a sufficient semantic neighborhood is established, adding more distant experiences provides no additional insights while diluting the prompt's context.

\begin{figure}[t]
    \centering
    \includegraphics[width= \linewidth]{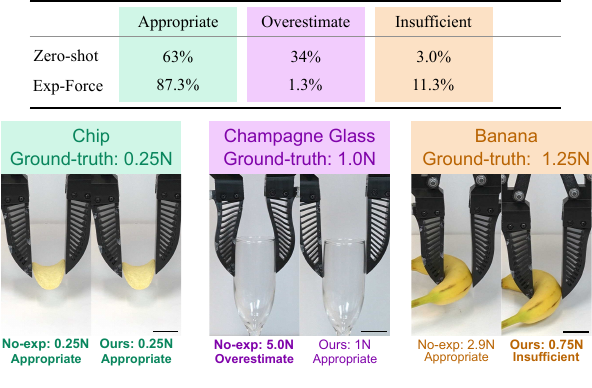}
    \caption{\textbf{Snapshots of representative trials}. We compare grasp outcomes of a Pringle chip, a champagne glass, and a banana, using forces generated by both our zero-shot baseline and \ourmethod{} with Gemini-3-flash and k=7 from the same image. The chip grasp is representative of a typical successful grasp. The Champagne Glass and Banana correspond respectively to typical failure modes - the zero-experience model overestimating, and our method under-estimating. All scale bars are 5cm.}
    \label{fig:rollouts}
    \vspace{-15pt}
\end{figure}

\section{Real Robot Experiment Results}

\definecolor{LightGreen}{HTML}{cef4e6}
\definecolor{LightBlue}{HTML}{def1f7}

\begin{table*}[t]
    \centering
    \small
    \setlength{\tabcolsep}{2pt}
    \caption{Real-world grasping performance using Gemini-3-Flash}
    \label{tab:realworld_results}
    \begin{tabular}{l>{\columncolor{LightGreen}}ccc>{\columncolor{LightBlue}}c@{\hskip 12pt}>{\columncolor{LightGreen}}ccc>{\columncolor{LightBlue}}c}
        \toprule
        & \multicolumn{4}{c}{\textbf{Zero-shot Baseline}}
        & \multicolumn{4}{c}{\textbf{\ourmethod{}\ (k=7)}} \\
        \cmidrule(r){2-5}\cmidrule(l{12pt}){6-9}
        Category
            & Appr. [\%] & Overest. [\%] & Insuff. [\%] & MAE\,$\pm$\,STD [N]
            & Appr. [\%] & Overest. [\%] & Insuff. [\%] & MAE\,$\pm$\,STD [N] \\
        \cmidrule(r){1-5}\cmidrule(l{6pt}){6-9}
        Bottles
            & 64 & 32 & 4 & 3.8$\pm$2.9
            & 92~{(+44\%)} & 0 & 8 & 1.08$\pm$0.76~{(-71\%)} \\
        Cylinders
            & 60 & 32 & 8 & 5.6$\pm$3.3
            & 80~{(+33\%)} & 0 & 20 & 0.62$\pm$0.73~{(-89\%)} \\
        Cuboids
            & 64 & 36 & 0 & 5.0$\pm$5.0
            & 100~{(+56\%)} & 0 & 0 & 0.81$\pm$0.92~{(-84\%)} \\
        Fragile--Heavy
            & 56 & 44 & 0 & 3.1$\pm$0.95
            & 84~{(+50.0\%)} & 0 & 16 & 0.11$\pm$0.10~{(-97\%)} \\
        Fragile--Light
            & 80 & 20 & 0 & 0.47$\pm$0.81
            & 96~{(+20.0\%)} & 4 & 0 & 0.03$\pm$0.03~{(-94\%)} \\
        Odd Shapes
            & 56 & 40 & 4 & 6.0$\pm$9.8
            & 72~{(+29\%)} & 4 & 24 & 0.90$\pm$0.45~{(-85\%)} \\
        \cmidrule(r){1-5}\cmidrule(l{6pt}){6-9}
        \textbf{Overall}
            & \textbf{63} & \textbf{34} & \textbf{3} & \textbf{4.0$\pm$4.8}
            & \textbf{87.3~{(+38\%)}} & \textbf{1.3} & \textbf{11.3} & \textbf{0.59$\pm$0.68~{(-85\%)}} \\
        \bottomrule
    \end{tabular}
    \vspace{-15pt}
\end{table*}



















In real world grasping experiments, we observe that the experience-conditioned inference of \ourmethod{} outperforms zero-shot VLMs, improving the overall rate of Appropriate force selection from $63\%$ to $87\%$ (Fig.~\ref{fig:rollouts}, Tab.~\ref{tab:realworld_results}). A two-proportion z-test confirms that this improvement is statistically significant ($p < 10^{-6}$). In addition, the MAE decreases from 4.0 \si{\newton} to 0.59 \si{\newton}, representing an $85\%$ reduction in force prediction error. These results indicate that \ourmethod{} provides a practical and reliable solution for real-world pre-grasp force selection, maintaining strong cross-object generalization across diverse categories.

The main source of \ourmethod{}'s performance improvement is the elimination of Overestimation errors. For example, the Champagne Glass only requires 1.0\,\si{\newton}, but the zero-shot VLMs' estimates range from 3.6N to 7.0N due to identifying the glass as slippery - that is 7$\times$ ground-truth $F^\star$, and risks breaking the glass. \ourmethod{} does sometimes underestimate the force required, thus leading to new failures to stably grasp the object, as in the banana. However, this limitation is unlikely to affect real deployments. In practice, pre-grasp force estimation is typically paired with adaptive slip detection, which compensates for insufficient initial force. In our experiments, activating the slip-correction algorithm from \cite{Shang2026forte} after an Insufficient force was applied resulted in successful grasps for all Insufficient test cases. 

Table~\ref{tab:realworld_results} shows a detailed comparison of the VLMs performances by object category and the change in MAE. The performance gains with our method are most significant across the Fragile-Heavy, Cylinders, and Cuboids categories. This is because the low-friction surfaces of these objects consistently lead to the zero-shot VLMs overestimating $F^\star$ (Sec.~\ref{sec:experience_impact}). Surprisingly, the zero-shot VLMs performed quite well on the Fragile-light category, leading to minimal performacne gains when using \ourmethod{}. This is because the VLMs are able to identify that these objects are fragile and light-weight, and therefore, it chose to stay close to the preset minimum gripper force command of $0.25\,\si{\newton}$. (Sec.~\ref{section:experience_pool}). Light crumpling does still occur in the paper cup example, when $0.5\,\si{\newton}$ is applied (shown on website). These results show that \ourmethod{} can be a practical and reliable pre-grasp force selector for real-world robot settings.

\section{Discussion}

In this work, we introduced \ourmethod{}, an experience-conditioned framework for estimating pre-grasp force from a single RGB image with VLMs. Using only a small experience pool from one-hour real robot interaction, the method achieves accurate force estimation with a best-case error of $0.43\,\si{\newton}$ in offline experiments, approaching the $0.2\,\si{\newton}$ tactile sensor uncertainty. In real-world grasping experiments, \ourmethod{} improves the rate of appropriate force selection from $63\%$ to $87\%$. These results show that \ourmethod{} enables reliable and data-efficient pre-grasp force estimation across diverse objects without explicit physical modeling.

Despite promising results, several limitations remain. 
First, \ourmethod{} relies on a single RGB image and therefore cannot directly measure key physical properties such as object mass. Visually identical objects may require substantially different gripping forces (e.g., empty vs. full opaque bottles), introducing unavoidable estimation errors. Incorporating additional sensing modalities, such as multi-view perception or tactile force feedback, may help reduce this ambiguity.


Second, performance is bounded by the reasoning reliability of the underlying VLM. While the framework often retrieves semantically plausible reference experiences, the subsequent inference process may produce inconsistent or physically implausible force estimates. This reflects a broader limitation of current foundation models when applied to physical decision-making tasks.


Finally, the current implementation relies on large frontier VLMs, which introduces nontrivial computational overhead and limits real-time deployment. Future work may explore model compression or distillation to improve efficiency, and evaluate the framework at larger scales across more diverse objects, grasping poses, and gripper embodiments.

Despite these limitations, our results still demonstrate that in-context inference over a small set of relevant prior interactions enables accurate and generalizable pre-grasp force prediction. 


\renewcommand*{\bibfont}{\scriptsize}
\begingroup
\printbibliography
\endgroup

\end{document}